\documentclass[preprint,12pt]{elsarticle}

\usepackage{amssymb}

\usepackage{hyperref} 
\usepackage{graphicx} 
\usepackage{float}
\usepackage{amsmath,amssymb,amsfonts}
\usepackage{algorithmic}

\usepackage{array} 
\newcolumntype{L}[1]{>{\raggedright\arraybackslash}p{#1}} 
\usepackage{multirow}

\usepackage{graphicx}

\journal{Knowledge-Based Systems}

\begin{document}

\begin{frontmatter}



\title{Attribute-Aware Implicit Modality Alignment for Text Attribute Person Search}


\author{Xin Wang, Fangfang Liu, Zheng Li, Caili Guo}

\affiliation{
            addressline={Beijing University of Posts and Telecommunications}, 
            city={Beijing},
            postcode={100876}, 
            country={China}}

\begin{abstract}

Text attribute person search aims to find specific pedestrians through given textual attributes, which is very meaningful in the scene of searching for designated pedestrians through witness descriptions. The key challenge is the significant modality gap between textual attributes and images.
Previous methods focused on achieving explicit representation and alignment through unimodal pre-trained models. Nevertheless, the absence of inter-modality correspondence in these models may lead to distortions in the local information of intra-modality.
Moreover, these methods only considered the alignment of inter-modality and ignored the differences between different attribute categories.
To mitigate the above problems, we propose an Attribute-Aware Implicit Modality Alignment (AIMA) framework to learn the correspondence of local representations between textual attributes and images and combine global representation matching to narrow the modality gap.
Firstly, we introduce the CLIP model as the backbone and design prompt templates to transform attribute combinations into structured sentences. This facilitates the model's ability to better understand and match image details.
Next, we design a Masked Attribute Prediction (MAP) module that predicts the masked attributes after the interaction of image and masked textual attribute features through multi-modal interaction, thereby achieving implicit local relationship alignment.
Finally, we propose an Attribute-IoU Guided Intra-Modal Contrastive (A-IoU IMC) loss, aligning the distribution of different textual attributes in the embedding space with their IoU distribution, achieving better semantic arrangement.
Extensive experiments on the Market-1501 Attribute, PETA, and PA100K datasets show that the performance of our proposed method significantly surpasses the current state-of-the-art methods.

\end{abstract}

\begin{graphicalabstract}
\end{graphicalabstract}

\begin{highlights}
\item Research highlight 1
\item Research highlight 2
\end{highlights}


\end{frontmatter}



\section{Introduction}

\begin{figure}[htbp]
    \begin{minipage}{\textwidth}
        \centering
        \includegraphics[width=0.8\textwidth]{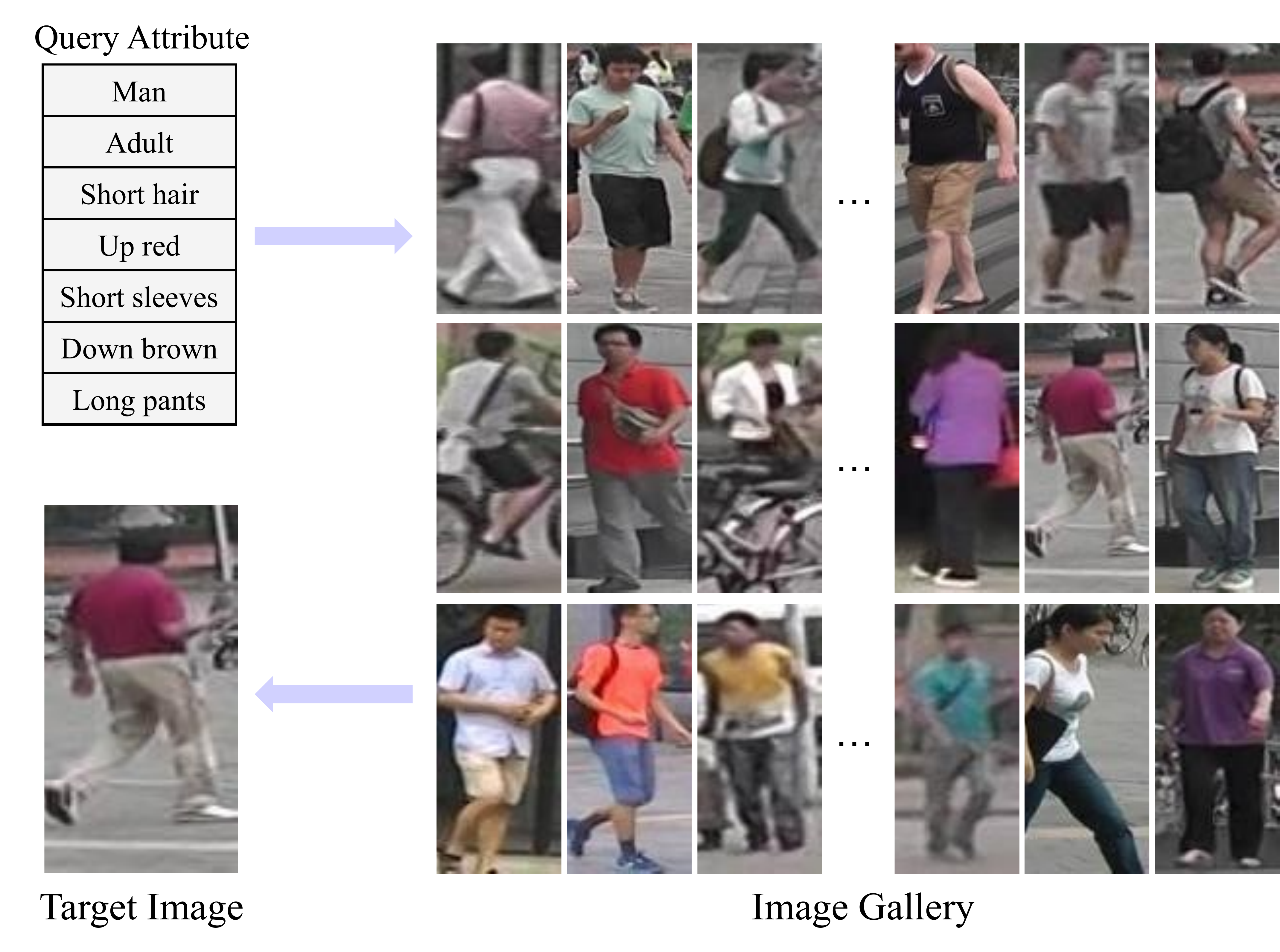}
    \end{minipage}
    \caption{Illustration of the text attribute person search task, which aims to identify the images which match the attribute query.}
    \label{fig1:example}
\end{figure}

The purpose of person search is to find target pedestrian that meet the given criteria in a large database of images. It is of great significance in the context of intelligent security, such as searching for suspects or missing persons through surveillance cameras in security systems.
The most common approach involves using a person image from surveillance as a query~\cite{a1,a9,a11,a15,a16} to search for similar images. However, in practical scenarios, many places lack surveillance cameras, making it difficult to obtain image queries. In such cases, specific pedestrians can only be identified through eyewitness testimony.
Some other existing methods use rich natural language descriptions~\cite{d1,d2,d3,d4,irra} to achieve person search. However, the uncertainty and noise in textual descriptions make the task more complex. In contrast, attribute descriptions such as gender, age, and clothing are simpler and easier to obtain. This has led to widespread attention from the academic community to text attribute person search~\cite{b3,b4,b5,b7}, as illustrated in Fig.~\ref{fig1:example}.

However, its application in reality remains highly challenging. On one hand, there exists a significant semantic gap between visual and textual modalities. Textual attributes have lower dimensions compared to visual data and are often sparse, making it difficult to effectively align these two modalities. On the other hand, the amount of data for text attributes is smaller than that for images, and many text attribute samples have minor differences, making retrieval more difficult.

To address the aforementioned challenges, mainstream methods extract more discriminative image and text attribute representations. These are then aligned in a common space through cross-modal matching.
This alignment process can be divided into global-based and local-based matching methods. 
Global-based matching methods~\cite{b4,b5,b1,b6} involve extracting global representation of images and text. 
At the end of the model, cross-modal constraints are applied to achieve alignment. 
However, due to the lack of interaction between modalities, many discriminative local representations may not align well, limiting the search performance. 
Local-based matching methods~\cite{b3,b2,b_fdaa}, on the other hand, extract local representations in different ways and perform fine-grained matching. 
While this approach enhances discriminative capabilities to some extent, the explicit representation often introduces noise during the modeling process. 
Additionally, it introduces extra computational overhead, restricting its practical applicability in real-world scenarios.
Furthermore, existing methods often rely on unimodal pre-trained models, such as utilizing ResNet~\cite{ResNet} and MLP for extracting image and attribute representations separately. This lack of correspondence information between different modalities during training may lead to distortions in the local information of intra-modality.

In this paper, we propose an Attribute-Aware Implicit Modality Alignment (AIMA) framework, which learns the local representation correspondence between text attributes and images, and combines global representation matching to narrow the modality gap. First, we utilize the Contrastive Language-Image Pre-training (CLIP)~\cite{clip} model as the backbone, as it has been trained on a large number of image-text pairs, encompassing rich multimodal information. At the same time, we design prompt templates to convert attribute combinations into structured sentences as input, enabling CLIP to better understand and match image details. By fine-tuning CLIP, its rich knowledge can be transferred to person search tasks, achieving improved multimodal representation alignment effects.

Secondly, we introduce a Masked Attribute Prediction (MAP) module that utilizes attention mechanism to facilitate multimodal information interaction between image and text attribute representations. Specifically, the interaction and fusion of image and masked text attribute representations are achieved through a multimodal encoder, followed by attribute prediction on the masked attributes. This enables implicit alignment of inter-modal and intra-modal representations. 
Analogous to the commonly used Masked Language Modeling (MLM)~\cite{MLM-Taylor} approach in natural language processing, this method allows the model to better understand fine-grained semantic correlations between text attributes and images, thereby learning more discriminative representations.

Lastly, existing methods often overlook differences between attributes, leading to embeddings of different attribute categories being close in the embedding space. The ASMR loss~\cite{b5} controls the distance between attributes in the embedding space based on their similarities, but it requires data from all attribute categories during training and cannot be trained in batches, making it impractical for large-scale datasets. We propose an Attribute-IoU Guided Intra-Modal Contrastive (A-IoU IMC) loss, using IoU between different attributes as similarity to adaptively align their distribution in the embedding space, thus clearly separating categories of different individuals for improved semantic organization.
The main contributions of this paper are as follows:
\begin{itemize}
\item We propose an Attribute-Aware Implicit Modal Alignment (AIMA) framework. We introduce the CLIP model for the text attribute person search task and design appropriate prompt templates to enable CLIP to better understand and match image details.

\item We design a Mask Attribute Prediction (MAP) task to implicitly capture the semantic relationship between textual attributes and images, leading to the acquisition of more discriminative representations.

\item We propose an Attribute-IoU Guided Intra-Modal Contrastive (A-IoU IMC) loss, which adaptively aligns with the attribute representation distribution in the embedding space to achieve better semantic arrangement.

\item We found that excellent results can be achieved just by fine-tuning CLIP, and experiments on the Market-1501 Attribute, PETA, and PA100K datasets also show that AIMA substantially outperforms current SOTA methods.
\end{itemize}


\section{Related Works}
\subsection{Text Attribute Person Search}
Text attribute person search aims to find corresponding pedestrians from an image database by giving attribute descriptions. It is mainly divided into methods based on attribute recognition~\cite{b8,b9,b10}, adversarial learning~\cite{b4,b1}, and cross-modal matching~\cite{b3,b5,b2,b_fdaa}. The attribute recognition methods first predict the attributes of the image, then calculate the similarity between the given attribute query and the predicted attributes to determine the target image. However, due to the low accuracy of attribute recognition, such methods are difficult to apply in practice.

Adversarial learning methods align the distributions of images and attributes in a common embedding space. Zhou et al.~\cite{b1} align the concepts generated by attributes with the concepts extracted from images. Cao et al.~\cite{b4} further improve performance through symbiotic adversarial learning that synthesizes unseen categories and optimizes the alignment of embeddings. However, this minimax optimization method suffers from training instability and difficulty in convergence.

Cross-modal matching methods align images and attributes in a common embedding space and make attributes close to their corresponding images. Dong et al.~\cite{b2} first define this problem as a zero-shot retrieval problem and perform global and local matching through hierarchical embeddings. However, it is computationally expensive and requires additional matching networks. Iodice et al.~\cite{b3} obtain local representations by dividing the image's global representations used for classification, but these local representations do not correspond well to specific parts of the body, leading to lower reliability. Jeong et al.~\cite{b5} control the distance between different attributes in the embedding space adaptively by considering the semantic relationships between them, achieving excellent performance on unseen categories. However, it only utilizes global representation, lacking fine-grained feature representations and having poor discriminability. Peng et al.~\cite{b_tip} address attribute labeling problems in the dataset by changing the distribution of original text attribute data through weak semantic embeddings, improving the representativeness of attribute representations.

All the above works use independently unimodal pre-trained models to extract features and then perform cross-modal alignment. This results in a lack of corresponding information across modalities, while vision-language pre-training models contain rich multimodal prior information. Jiang et al.~\cite{irra} found that simply fine-tuning the CLIP~\cite{clip} model can effectively transfer its knowledge to text-image person retrieval, achieving excellent performance.
The proposed AIMA in this work can more effectively transfer the powerful prior knowledge of CLIP to text attribute person search, reducing the semantic gap between modalities.

\subsection{Vision-Language Pre-training}
Vision-Language Pre-training (VLP)~\cite{uniter,empirical,align,vilbert} refers to the process of pre-training on a large-scale dataset of image-text pairs to learn the semantic correspondence between different modalities. Inspired by pre-trained language models and the increasing trend of using transformer-based~\cite{transformer} architectures in both the Natural Language Processing (NLP) and Computer Vision (CV) domains, VLP has become the mainstream training paradigm for solving vision-language tasks. It has demonstrated robust performance in downstream tasks such as visual question answering~\cite{visual-question-answer}, image captioning~\cite{image-caption}, visual reasoning~\cite{visual-reasoning}, and image-text retrieval~\cite{image-text-retrieval}.

VLP models typically extract visual features and text features from image-text pairs separately using visual encoders and text encoders, respectively. These features are then fed into a multimodal fusion module to obtain cross-modal representations. Based on different multimodal fusion approaches, VLP works can be categorized into single-stream~\cite{uniter,vilt} and dual-stream~\cite{clip,empirical,scalingUp} structures. The single-stream structure combines the encoded features of vision and text together and inputs them into a single transformer block, merging multimodal inputs through merged attention. Due to sharing the same parameter set for both modalities, the single-stream structure is more parameter-efficient. However, it needs to predict similarity scores for all possible image-text pairs during inference, leading to slower retrieval speed. In contrast, in the dual-stream structure, the encoded features of vision and text are not combined but independently inputted into two separate transformer blocks without parameter sharing, achieving cross-modal interaction through co-attention. Since it only needs to encode image and text features once during inference, its retrieval performance is higher. However, it lacks the ability to model complex cross-modal interactions in other visual-language understanding tasks.

The pre-training objectives of VLP can be aptly delineated into four primary categories: Masked Language Modeling (MLM)~\cite{MLM-Taylor}, Masked Image Modeling (MIM)~\cite{MIM}, Vision-Language Matching (VLM)~\cite{align}, and Vision-Language Contrastive Learning (VLC)~\cite{clip}. Originating from the realm of NLP, MLM endeavors to predict masked text tokens through the utilization of unmasked tokens, while MLM within the VLP paradigm augments this approach by incorporating visual tokens for prediction. Similarly, MIM within VLP orchestrates the reconstruction of masked patches by amalgamating visible image patches and all text tokens. VLM treats the matching of visual and text features as a binary classification problem, predicting whether they match based on the fused representations of both modalities. VLC, which aligns visual and textual features before performing modal fusion, emerges as a popular pre-training objective. Recent text-image person retrieval works \cite{irra,rasa} utilize MLM as the fine-tuning objective, which effectively transfers the prior knowledge of pre-trained models and achieves fine-grained representation matching.


\section{Method}

\begin{figure*}[t!]
    \centering
    \includegraphics[width=\textwidth]{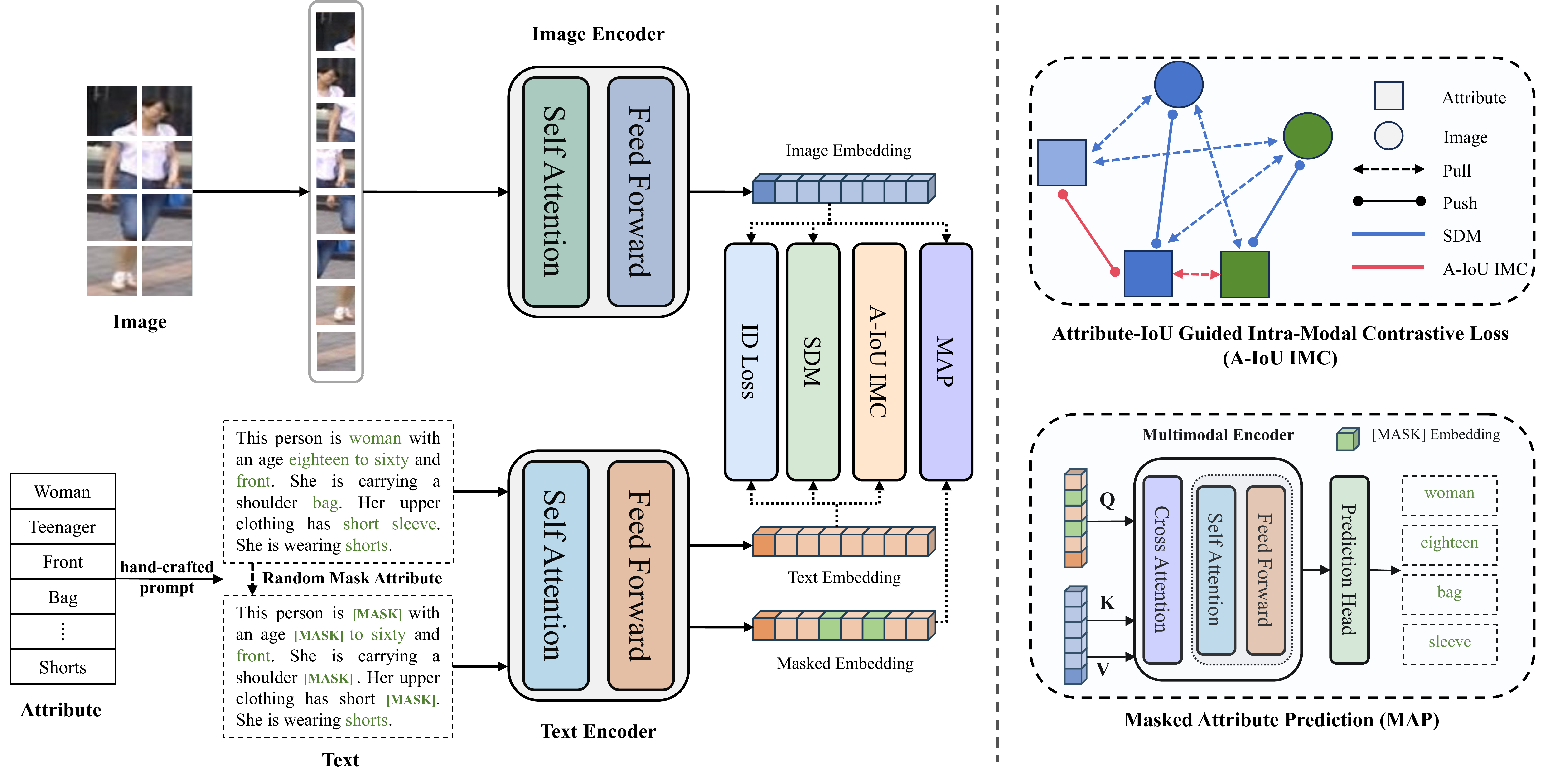}
    \caption{
    The overview of our proposed AIMA framework. It consists of an image encoder, a text encoder, and a multimodal encoder. It has four representation learning branches: Masked Attribute Prediction (MAP), Attribute-IoU Guided Intra-Modal Contrastive loss (A-IoU IMC), Similarity Distribution Matching (SDM), and Identity loss (ID loss). AIMA achieves end-to-end training through these four tasks, requiring only the computation of global feature representations during inference.
    }
    \label{fig2:framework}
    \vspace{-2mm}
\end{figure*}

In this section, we will explain our proposed Attribute-Aware Implicit Modality Alignment (AIMA) framework for text attribute person search in Fig.~\ref{fig2:framework}. All the details are elaborated upon in the subsequent subsections.

\subsection{Feature Extraction Encoder} 
In prior works on text attribute person search, feature extraction primarily relies on unimodal pre-trained models. ResNet~\cite{ResNet} is typically used for images, while MLP is common for text attributes. However, these models lack cross-modal correspondence information inherently. Recent work~\cite{irra} inspires us to transfer CLIP's knowledge to cross-modal person retrieval tasks. Therefore, we also use the pre-trained CLIP model to initialize the encoders in AIMA. This allows for the effective transfer of its powerful prior knowledge through fine-tuning.

\subsubsection{Image Encoder}
Given an input image $I \in \mathbb{R}^{H\times W\times C}$, the CLIP pre-trained Vision Transformer (ViT)~\cite{vit} model serves as the image encoder to produce the image embedding. The process initiates by partitioning the input image into $N$ non-overlapping patches of fixed dimensions, where $N$ is computed as $\frac{H \times W}{P^2}$, with $P$ denoting the patch size. Consequently, these patches are transformed into a sequence of one-dimensional tokens $\{f_1^v, f_2^v,..., f_N^v\}$ through a trainable linear projection layer. To preserve spatial relationships among the patches, an additional [CLS] token is introduced, followed by positional encoding being applied to the sequence, thereby constructing the model's final input as $\{f_{cls}^v, f_1^v,..., f_N^v\}$. This augmented sequence then traverses through a series of L-layer transformer blocks, leveraging self-attention mechanisms to distill the information and yield a refined feature representation. Ultimately, from this feature representation, the token $f_{cls}^v$ is linearly transformed to derive the global feature, which acts as a pivotal component in facilitating image-text matching.

\subsubsection{Text Encoder}
\label{sec:text_encoder}
Given a set of attribute descriptions $A$, such as \{woman, adult, bag,..., shorts\}, we initially transform them into corresponding sentence representations $T$ via hand-crafted prompt templates. The templates used across three datasets are illustrated in Tab.~\ref{tab:prompt}, where slotting the corresponding attribute terms into their respective positions yields a sentence describing a pedestrian.

The motivation behind transforming attribute descriptions into corresponding sentences as inputs to a text encoder can be attributed to several reasons: 1) The training data for the CLIP model employs sentences rather than individual words or phrases. 2) Sentences offer a more substantial context, facilitating the model's understanding and matching of image details. 3) Utilizing sentences as inputs aligns more closely with human linguistic habits, thereby enhancing the model's performance in practical applications.

\begin{table*}[ht!]
    \centering
    \resizebox{\textwidth}{!}{
    \begin{tabular}{ l | c }
    \hline
        Dataset & Prompt Templates \\ \hline
        \multirow{3}{*}{Market-1501~\cite{Market1501}} 
        &  A [Age] [Gender] has [Hair length] hair. [Gender] carries a [Bag]. [Gender] upper body \\
        & is [Upper color] with [Sleeve length] sleeve. [Gender] lower body is [Lower color] with  \\
        & [Lower clothing length] [Lower clothing type]. [Gender] wears a [Hat].
        \\
        \hline

        \multirow{5}{*}{PETA~\cite{PETA}} 
        & A [Gender] with an age [Age] carries [Carrying]. [Gender] has lower-[Kind of lower body], upper- \\
        & [Kind of uppper body] style and [Sleeve]. [Gender] has [Hair color] [Hair length] hair. [Gender] accessories \\
        & include [Accessory]. [Gender] upper clothes are [Upper color] [Upper clothing], [Upper texture]. [Gender] lower \\
        & clothes are [Lower color] [Lower clothing], [Lower texture].[Gender] wears [Footwear color] [Footwear].\\  
        \hline
        
        \multirow{4}{*}{PA100K~\cite{PA100K}} 
        & A [Gender] with an age [Age] stands [Viewpoint]. [Gender] wears a [Hat] and [Glasses]. [Gender] carries a\\ 
        & [Bag]. [Gender] hold objects in front. [Gender] upper clothing has [Sleeve length] sleeve and [Upper patterns]. \\
        & [Gender] lower clothing has [Lower patterns]. [Gender] wears [Lower clothing] and footwear are [Boots]. \\
        \hline

    \end{tabular}}
    \vspace{0.1mm}
    \caption{Prompt templates for transforming attribute descriptions into sentence representations.}
    \vspace{-2mm}
    \label{tab:prompt}
\end{table*}

For the text sentence $T$, we employ CLIP's text encoder, which is an enhanced Transformer~\cite{clip}, to perform text feature extraction. Specifically, the input text is first tokenized using lower-case Byte Pair Encoding (BPE)~\cite{vocab_size} with a vocabulary size of 49,152. The sequence of tokens resulting from this process is then prefixed and suffixed with [SOS] and [EOS] tokens respectively, to denote the start and end of the sentence. Each token is subsequently transformed into an embedding representation, and position encoding is added to form the final input sequence $\{f_{sos}^t, f_1^t,..., f_{eos}^t\}$. This sequence is then fed into the transformer, where self-attention mechanisms are employed to learn correlations between different tokens. Ultimately, the feature representation $f_{eos}^t$ at the end-of-sentence position from the last layer of the transformer is projected through a linear layer to derive the global feature for the corresponding text.

\subsection{Masked Attribute Prediction}

One of the major challenges in text attribute person search is the modality gap between visual and textual information. Effectively extracting and representing their fine-grained features can help mitigate this problem. Previous approaches~\cite{b3,b2,b_fdaa} have primarily focused on explicitly depicting local features and aligning them. However, this approach often introduces noise and increases computational costs during the modeling process. To address these limitations, this paper introduces the Masked Attribute Prediction (MAP) module, which aims to achieve implicit modality alignment. Specifically, we employ Masked Attribute Modeling (MAM) to predict masked attributes. By utilizing MAM, we can implicitly model the fine-grained relationship between images and texts, enabling the model to learn more discriminative global features.

\subsubsection{Masked Attribute Modeling}

The Masked Attribute Modeling (MAM) introduced in this paper is inspired by the widely used Masked Language Modeling (MLM)~\cite{MLM-Taylor} in the field of Natural Language Processing (NLP). The main idea of MLM is to learn high-quality text representations by predicting the correct forms of words that are randomly masked in the input text. When applied to cross-modal tasks, it requires a combination of unmasked text tokens and visual tokens to predict the masked text tokens. The text attribute person search task often requires the model to focus more on specific attribute words in the sentence. By randomly masking and predicting only the attribute words describing a person in the sentence, a more fine-grained learning objective is provided. This approach facilitates the model's better understanding of key parts of the text while promoting the correspondence between local image regions and textual attributes. In this paper, we use a multimodal encoder to predict the masked attribute words by using text tokens with randomly masked attribute words and visual tokens. It aligns image and text contextual representations with embeddings of the masked text tokens and can implicitly use fine-grained local information to achieve global feature alignment.

\subsubsection{Multimodal Encoder}
The multimodal encoder facilitates better fusion of image and text embeddings and achieves implicit alignment through attribute prediction. We adopt an efficient multimodal encoder designed by Jiang et al.~\cite{irra}, as shown in Fig.~\ref{fig2:framework}. This multimodal encoder consists of a multi-head cross-attention (MCA) layer and 4-layer transformer blocks, which are structurally simple and computationally efficient. 

Specifically, given an input attribute description $A$, it is converted into a text sentence $T$ through the corresponding prompt template. Since we know the positions of specific attribute words in the sentence, we randomly mask the attribute tokens in the sentence with a probability of 15\% and replace them with a special token [MASK]. Unlike the masking strategy in BERT~\cite{bert}, the replacements are 10\% remain unchanged, and 90\% become [MASK]. This avoids introducing additional noise in the text that may affect the interaction between images and text. The masked text $\hat{T}$ is input into the text encoder in Sec.~\ref{sec:text_encoder}. The final hidden states of the text encoder and the image encoder, $\{h_{i}^{\hat{t}}\}_{i=1}^{L}$ and $\{h_{i}^{v}\}_{i=1}^{N}$ respectively, are input into the multimodal encoder. $h_{i}^{\hat{t}}$ represents the masked text representation, where $L$ denotes the length of text tokens. $h_{i}^{v}$ represents the image representation, where $N$ denotes the number of image patches. Among them, $\{h_{i}^{\hat{t}}\}_{i=1}^{L}$ serves as the query ($\mathcal{Q}$) for MCA, and $\{h_{i}^{v}\}_{i=1}^{N}$  serves as the key ($\mathcal{K}$) and value ($\mathcal{V}$). Therefore, the fusion process of image and masked text representations can be represented as:
\begin{equation}
\{h_{i}^{f}\}_{i=1}^{L} = Transformer(MCA(LN(\mathcal{Q},\mathcal{K},\mathcal{V}))),
\label{eq:multimodal_encoder}
\vspace{-1mm}
\end{equation}
where $\{h_{i}^{f}\}_{i=1}^{L}$ represents the fused representation, $LN(\cdot)$ denotes layer normalization, and the calculation of $MCA(\cdot)$ is as follows:
\begin{equation}
MCA(\mathcal{Q},\mathcal{K},\mathcal{V}) = softmax \left(\frac{\mathcal{Q}\mathcal{K}^{T}}{\sqrt{d}}\right)\mathcal{V},
\label{eq:MCA}
\vspace{-1mm}
\end{equation}
where $d$ represents the dimension of masked text embedding.

Since we need to predict for each masked position $h_{i}^{f}$, we use an MLP as the prediction head to predict the probability of the corresponding original attribute token. The following cross-entropy loss is adopted as the optimization objective:
\begin{equation}
\mathcal{L}_{map} = -\frac{1}{|\mathcal{F}||\mathcal{G}|} \sum\limits_{i\in \mathcal{F}} \sum\limits_{j\in \mathcal{G}} y_j^{i} \log \frac{\exp({r_j^{i}})} {\sum_{k=1}^{|\mathcal{G}|} \exp({r_k^{i}}) },
\label{eq:loss_map}
\vspace{-1mm}
\end{equation}
where $\mathcal{F}$ represents the set of masked text tokens, $|\mathcal{G}|$ represents the size of the vocabulary $\mathcal{G}$, $r^i$ represents the probability distribution of the predicted token, and $y^i$ represents the one-hot distribution of the true labels.

\subsection{Attribute-IoU Guided Intra-Modal Contrastive Loss}

Previous methods often overlook differences between attributes, leading to embeddings of different attribute categories being close in the embedding space. This makes the distances between the corresponding image embeddings closer, reducing the accuracy of retrieval results. We propose a novel Attribute-IoU Guided Intra-Modal Contrastive (A-IoU IMC) loss, using IoU between different attributes as similarity to adaptively align their text distribution in the embedding space, thus clearly separating categories of different individuals for improved semantic organization. An intuitive example is shown in Fig.~\ref{fig5:embedding_vector}.

Specifically, given a batch of image-attribute pairs with a size of $B$, we can obtain the corresponding image-text pairs through the prompt template. For each global text representation $f_i^t$, we construct a set of text-text representation pairs $\{ f_i^t, f_j^t \}_{j=1}^B$. Let $s(\mathbf{u}, \mathbf{v})=\mathbf{u}^T \mathbf{v} / \left\|\mathbf{u}\right\|\left\|\mathbf{v}\right\| $ denote the cosine similarity between $\mathbf{u}$ and $\mathbf{v}$. Then, the matching probability between text pairs can be computed using the following function:

\begin{equation}
p_{i,j} = \frac{\exp(s(f_i^t,~f_j^t))}{\sum_{k=1}^{B} \exp(s(f_i^t,~f_k^t))}.
\label{eq:text_distribution}
\vspace{-1mm}
\end{equation}
The matching probability $p_{i,j}$ can be viewed as the ratio of the cosine similarity score between $f_i^t$ and $f_j^t$, and the sum of similarity scores between $f_i^t$ and all other $f_k^t$ in the batch.

\begin{figure*}[htbp]
    \centering
    \includegraphics[width=\textwidth]{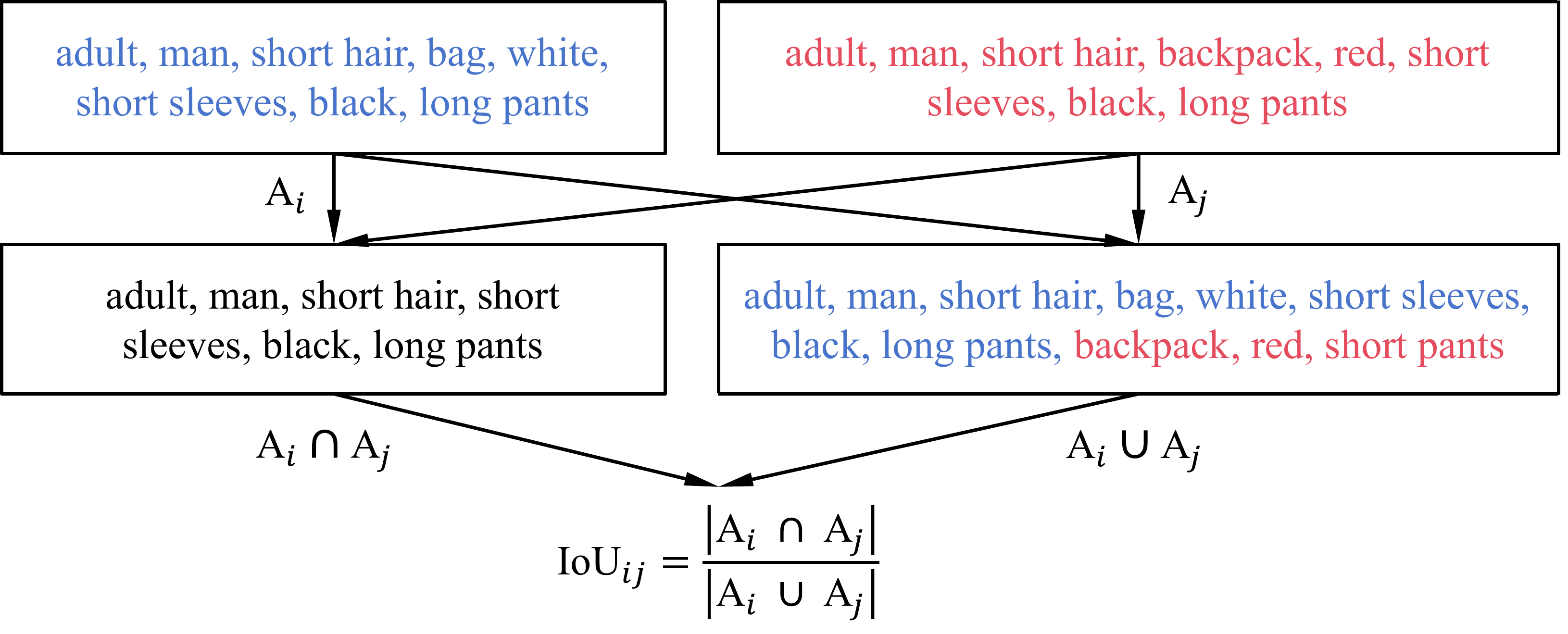}
    \caption{Illustration of the calculation process of Attribute-IoU.}
    \label{fig3:iou_example}
    \vspace{-2mm}
\end{figure*}

To facilitate a better semantic arrangement of texts belonging to different categories in the embedding space, we adopt the IoU of the corresponding attributes as soft-labels for supervision. An example of Attribute-IoU is depicted in Fig.~\ref{fig3:iou_example}, which can be defined as follows:
\begin{equation}
IoU_{i,j} = \frac{|A_i~\cap~A_j|}{|A_i~\cup~A_j|},
\label{eq:IoU}
\vspace{-1mm}
\end{equation}

where $|A_i \cap A_j|$ denotes the count of common attribute words, and $|A_i \cup A_j|$ represents the total number of attribute words in the union set. $IoU=1$ indicates that the two attributes are identical, whereas $IoU=0$ signifies that the two attributes have no common words.

The Attribute-IoU serves as soft-labels for the similarity between text pairs and can be computed by:
\begin{equation}
q_{i,j} = \frac{IoU_{i,j}}{\sum_{k=1}^{B} IoU_{i,k}}.
\label{eq:IoU_distribution}
\vspace{-1mm}
\end{equation}

The A-IoU IMC loss is defined as the cross-entropy loss between the true matching distribution $q_{i,j}$ between text pairs and the matching probability $p_{i,j}$. Ultimately, for each batch, the A-IoU IMC loss can be computed by:
\begin{equation}
\mathcal{L}_{IoU} = -\frac{1}{B} \sum_{i=1}^{B} \sum_{j=1}^{B} q_{i,j} \log p_{i,j}.
\label{eq:loss_iou}
\vspace{-1mm}
\end{equation}

\subsection{Overall Loss}

The optimization objective of AIMA aims to improve the learning of global feature representations of images and text in the embedding space. Therefore, in addition to the aforementioned losses, we also employ the SDM loss~\cite{irra} and the ID loss~\cite{id_loss} as the optimization objectives for AIMA. The SDM loss aims to minimize the KL divergence between the distribution of image-text similarity and the normalized label matching distribution. It can reduce the distance between positive pairs and increase the distance between negative pairs in the embedding space, thus aligning the representations of image and text. The ID loss is a classification loss based on the identity of images and text. It can tightly cluster representations of images or text from the same identity in the embedding space. 

In summary, AIMA is trained in an end-to-end manner, and the final objective is as follows:
\begin{equation}
\mathcal{L} = \mathcal{L}_{map} + \mathcal{L}_{IoU} + \mathcal{L}_{sdm} + \mathcal{L}_{id}.
\label{eq:loss_all}
\vspace{-1mm}
\end{equation}


\section{Experiments}

In this section, we first describe the three datasets and settings for evaluation. Then we compare our approach with the state-of-the-art methods and conduct the ablation studies. Finally, we show the qualitative results of our method.

\subsection{Datasets and Evaluations}

We extensively evaluate our proposed AIMA framework on three challenging
text attribute person search datasets.

\begin{table*}[ht!]
    \centering
    \resizebox{\textwidth}{!}{
    \begin{tabular}{ l | c }
    \hline
        Dataset & Attribute category \\ \hline

        \multirow{2}{*}{Market-1501~\cite{Market1501}} & Age, Bag, Color of lower body clothing, Color of upper body clothing, Type of lower body clothing, \\
        &Length of lower body clothing, Sleeve length, Hair length, Hat, Gender \\ \hline 
        
        \multirow{3}{*}{PETA~\cite{PETA}} & Age, Carrying, Upper body casual, Lower body casual, Accessory, Footwear,\\ 
        & Kind of upper body, Sleeve, Kind of lower body, Texture of upper body, Texture of lower body,\\
        & Gender, Hair length, Color of upper body, Color of lower body, Color of footwear, Color of hair \\
        \hline
        
        \multirow{2}{*}{PA100K~\cite{PA100K}} & Age, Gender, Viewpoint, HandBag, ShoulderBag, Backpack, HoldObjectsInFront, Hat, Glasses\\
        &  Length of Sleeve, Patterns of upper body, Patterns of lower body, Coat, Kind of lower body, Boots\\
        \hline
    \end{tabular}}
    \vspace{0.1mm}
    \caption{Lists of attribute categories in the three benchmark datasets for text attribute person search.
    }
    \vspace{-2mm}
    \label{tab:attribute-groups}
\end{table*}

\textbf{Market-1501 Attribute}~\cite{Market1501} is an extension of the person re-identification benchmark Market-1501, which contains 12,936 training images and 16,483 images for testing. The training set consists of 508 identities, and the test set consists of 484 identities. The resolution of images is $128 \times 64$, each image is labeled with 27 different attribute words. There are 10 categories of attribute descriptions, as shown in Tab.~\ref{tab:attribute-groups}.

\textbf{PETA}~\cite{PETA} is constructed by 10 public small person re-identification datasets with various resolutions, ranging from $17 \times 39$ to $169 \times 365$. The dataset contains 12,140 training images and 1,181 testing images. The training set consists of 1,890 identities, and the test set consists of 200 identities. Each image is labeled with 65 different attribute words.

\textbf{PA100K}~\cite{PA100K} is a recently introduced large-scale pedestrian attribute dataset. The dataset contains 80,000 training images and 10,000 testing images. The training set consists of 2,020 identities, and the test set consists of 849 identities. Each image is labeled with 26 commonly used attribute words. PA100K is more challenging than Market-1501 Attribute and PETA.

\textbf{Evaluation Metrics.}
Similar to previous works, we use the popular Rank-k metrics (k=1,5,10) and mean average precision (mAP) as evaluation metrics. Rank-k represents the probability of finding at least one matching image among the top-k candidate images when given a text query. mAP is computed by averaging the area under the Precision-Recall curve across all test queries, thereby providing a more comprehensive evaluation. Higher values of Rank-k and mAP indicate better retrieval performance.

\subsection{Implemention Details}

The AIMA framework proposed in this paper consists of a pre-trained image encoder (CLIP-ViT-B/16), a pre-trained text encoder (CLIP Text Transformer), and a randomly initialized multimodal encoder. The embedding dimension of each layer in the multimodal encoder is 512, with 8 heads. During training, random horizontal flipping, random cropping with padding, and random erasing are used for image data augmentation, while random deletion is employed for text data augmentation. All input images are resized to $256 \times 128$, with a patch size of 16. The maximum length of the text-token sequence is 77. We use the Adam optimizer~\cite{adam} with an initial learning rate of $1e{-5}$ and cosine decay for training. The training epochs for Market-1501 Attribute, PA100K, and PETA are 10, 20, and 60, respectively. The batch size for training is 128. The learning rate is increased from $1e{-6}$ to $1e{-5}$ during the first 5 epochs. For the multimodal encoder, the initial learning rate is $5e{-5}$. During the inference stage, only the image encoder and text encoder are utilized. We perform our experiments on a single NVIDIA Tesla V100 GPU.

\begin{table*}[ht!]
    \centering
    \resizebox{\textwidth}{!}{
    \begin{tabular}{ l | c  c | c  c  c  c }
    \hline
        Method & Type & Ref & Rank-1 & Rank-5 & Rank-10 & mAP   \\ \hline

        DeepMAR~\cite{b8} & G & ACPR15 & 13.2 & 24.9 & 32.9 & 8.9  \\ 
        DCCAE~\cite{b11} & G & ICML15 & 8.1 & 24.0 & 34.6 & 9.7  \\ 
        2WayNet~\cite{b12} & G & CVPR17 & 11.3 & 24.4 & 31.5 & 7.8  \\ 
        CMCE~\cite{b13} & G & ICCV17 & 35.0 & 51.0 & 56.5 & 22.8  \\ 
        AAIPR~\cite{b1} & G & IJCAI18 & 40.3 & 49.2 & 58.6 & 20.7  \\ 
        AIHM~\cite{b2} & L & ICCV19 & 43.3 & 56.7 & 64.5 & 24.3  \\ 
        SAL~\cite{b4}$^\dagger$ & G & ECCV20 & 44.4 & 65.7 & 72.5 & 29.4  \\ 
        ASMR \cite{b5} & G & ICCV21 & 49.6 & 64.9 & 72.5 & 31.0  \\ 
        WSFG \cite{b_tip} & G & TIP23 & 47.5 & 63.2 & 73.1 & 29.0  \\ 
        GCN \cite{b_tnnls} & G & TNNLS23 & 48.8 & 67.2 & 78.0 & 30.2  \\ 
        GAN \cite{b_tmm} & G & TMM23 & 49.7 & 66.4 & 79.6 & 32.8  \\
        FDAA \cite{b_fdaa} & L & VCIP23 & 50.4 & 67.2 & 74.4 & 33.9  \\
        \hline
        \textbf{Baseline(CLIP-ViT-B/16)} & G & - & 47.5 & 69.0 & 76.0 & 36.7  \\
        \textbf{AIMA(Ours)} & G & - & \textbf{57.0} & \textbf{77.1} & \textbf{83.9} & \textbf{44.4}  \\ \hline

    \end{tabular}}
    \vspace{0.1mm}
    \caption{Performance comparisons with state-of-the-art methods on Market-1501 Attribute dataset. ``G" and ``L" in ``Type" column stand for global-matching/local-matching method. $\dagger$ indicates results reproduced by the official implementation.
    }
    \vspace{-2mm}
    \label{tab:results on market}
\end{table*}

\begin{table*}[ht!]
    \centering
    \resizebox{\textwidth}{!}{
    \begin{tabular}{ l | c  c | c  c  c  c }
    \hline
        Method & Type & Ref & Rank-1 & Rank-5 & Rank-10 & mAP   \\ \hline

        DeepMAR~\cite{b8} & G & ACPR15 & 17.8 & 25.6 & 31.1 & 12.7  \\ 
        DCCAE~\cite{b11} & G & ICML15 & 14.2 & 22.1 & 30.0 & 14.5  \\ 
        2WayNet~\cite{b12} & G & CVPR17 & 23.7 & 38.5 & 41.9 & 15.4  \\ 
        CMCE~\cite{b13} & G & ICCV17 & 31.7 & 39.2 & 48.4 & 26.2  \\ 
        AAIPR~\cite{b1} & G & IJCAI18 & 39.0 & 53.6 & 62.2 & 27.9  \\ 
        SAL~\cite{b4}$^\dagger$ & G & ECCV20 & 39.0 & 61.5 & 70.0 & 37.2  \\ 
        ASMR \cite{b5} & G & ICCV21 & 56.5 & 80.0 & 83.5 & 50.2  \\ 
        WSFG \cite{b_tip} & G & TIP23 & 59.0 & 81.5 & 86.5 & 53.9  \\ 
        GCN \cite{b_tnnls} & G & TNNLS23 & 49.1 & 68.8 & 77.0 & 42.6  \\ 
        GAN \cite{b_tmm} & G & TMM23 & 56.9 & 79.4 & 84.1 & 52.4  \\
        FDAA \cite{b_fdaa} & L & VCIP23 & 60.0 & 86.0 & 90.5 & 51.7  \\
        \hline
        \textbf{Baseline(CLIP-ViT-B/16)} & G & - & 73.5 & 85.0 & 91.0 & 68.0  \\
        \textbf{AIMA(Ours)} & G & - & \textbf{83.0} & \textbf{96.0} & \textbf{98.0} & \textbf{78.1}  \\ \hline

    \end{tabular}}
    \vspace{0.1mm}
    \caption{Performance comparisons with state-of-the-art methods on PETA dataset.}
    \vspace{-2mm}
    \label{tab:results on peta}
\end{table*}

\begin{table*}[ht!]
    \centering
    \resizebox{\textwidth}{!}{
    \begin{tabular}{ l | c  c | c  c  c  c }
    \hline
        Method & Type & Ref & Rank-1 & Rank-5 & Rank-10 & mAP   \\ \hline

        DCCAE~\cite{b11} & G & ICML15 & 21.2 & 39.7 & 48.0 & 15.6  \\ 
        2WayNet~\cite{b12} & G & CVPR17 & 19.5 & 26.6 & 34.5 & 10.6\\ 
        CMCE~\cite{b13} & G & ICCV17 & 25.8 & 34.9 & 45.4 & 13.1  \\ 
        AIHM~\cite{b2} & L & ICCV19 & 31.3 & 45.1 & 51.0 & 17.0  \\ 
        SAL~\cite{b4}$^\dagger$ & G & ECCV20 & 22.7 & 36.5 & 41.6 & 15.0  \\ 
        ASMR \cite{b5} & G & ICCV21 & 31.9 & 49.1 & 58.2 & 20.6  \\ 
        WSFG \cite{b_tip} & G & TIP23 & 30.2 & 49.7 & 58.8 & 22.4  \\ 
        FDAA \cite{b_fdaa} & L & VCIP23 & 33.3 & 53.4 & 61.0 & 23.6  \\
        \hline
        \textbf{Baseline(CLIP-ViT-B/16)} & G & - & 35.7 & 59.8 & 70.4 & 32.2  \\
        \textbf{AIMA(Ours)} & G & - & \textbf{42.5} & \textbf{63.7} & \textbf{72.1} & \textbf{36.3}  \\ \hline

    \end{tabular}}
    \vspace{0.1mm}
    \caption{Performance comparisons with state-of-the-art methods on PA100K dataset.}
    \vspace{-2mm}
    \label{tab:results on pa100k}
\end{table*}

\subsection{Comparisons with the State-of-the-Arts}

In this subsection, we present the comparison results with state-of-the-art methods on three benchmark datasets. The Baseline method in the table refers to the CLIP model fine-tuned solely with the InfoNCE loss~\cite{infonce}, without any additional components.

\textbf{Performance Comparisons on Market-1501 Attribute.} 
We first evaluate the proposed method on the widely used Market-1501 Attribute dataset. As shown in Tab.~\ref{tab:results on market}, AIMA achieves a Rank-1 accuracy of 57.0\% and an mAP of 44.4\%, surpassing the state-of-the-art method by 13.1\% and 31.0\%, respectively. Furthermore, it is noteworthy that the Baseline, established through direct fine-tuning of CLIP, already demonstrates comparable performance to the  state-of-the-art methods, achieving a Rank-1 accuracy of 47.5\% and an mAP of 36.7\%. Previous methods mainly employed ResNet as the backbone network, while CLIP's backbone network is ViT, which has powerful feature extraction capabilities. CLIP also possesses rich multimodal prior knowledge, which can be effectively transferred to cross-modal tasks to improve performance.

\textbf{Performance Comparisons on PETA.} 
We also evaluate the proposed method on the PETA dataset. As shown in Tab.~\ref{tab:results on peta}, AIMA achieves a Rank-1 accuracy of 83.0\% and an mAP of 78.1\%, representing a substantial improvement over previous state-of-the-art methods by 38.3\% and 44.9\%, respectively. Additionally, the Baseline achieved by directly fine-tuning CLIP also surpasses the state-of-the-art methods considerably, achieving a Rank-1 accuracy of 73.5\% and an mAP of 68.0\%. This further demonstrates CLIP's powerful feature extraction and transfer learning capabilities.

\textbf{Performance Comparisons on PA100K.} 
Experimental results on the PA100K dataset are presented in Tab.~\ref{tab:results on pa100k}. AIMA achieves a Rank-1 accuracy of 42.5\% and an mAP of 36.3\%, surpassing the state-of-the-art methods by 27.6\% and 53.8\% respectively. Meanwhile, the Baseline obtained by directly fine-tuning CLIP also significantly outperforms the most advanced methods, achieving a Rank-1 accuracy of 35.7\% and an mAP of 32.2\%.

In summary, the proposed AIMA achieves significant improvements over state-of-the-art methods on three datasets. This indicates its ability to effectively transfer the prior knowledge of CLIP to the person search task, exhibiting both discriminative and generalization capabilities.

\subsection{Ablation Studies}

In this subsection, we analyze the effectiveness of the proposed components in the AIMA framework. The Baseline model refers to the CLIP-ViT-B/16 model fine-tuned with the InfoNCE loss, without any additional components. All results are reported in Tab.~\ref{tab:ablation results}.

\begin{table*}[ht!]
    \centering
    \resizebox{\textwidth}{!}{
\begin{tabular}{c | l | c c c c | c c c c | c c c c}
\hline
    {\multirow{2}{*}{No.}} & \multirow{2}{*}{Methods} & \multicolumn{4}{c|}{Market-1501 Attribute} & \multicolumn{4}{c|}{PETA} & \multicolumn{4}{c}{PA100K}  \\ \cline{3-14}
   &   & Rank-1 & Rank-5 & Rank-10 & mAP & Rank-1 & Rank-5 & Rank-10 & mAP & Rank-1 & Rank-5 & Rank-10 & mAP  \\  \hline
   0 & Baseline                                    & 47.5 & 69.0 & 76.0 & 36.7 & 73.5 & 85.0 & 91.0 & 68.0 & 35.7 & 59.8 & 70.4 & 32.2  \\
   1 & $+$prompt                                   & 50.4 & 70.5 & 76.2 & 39.8 & 78.5 & 92.5 & 95.5 & 74.9 & 37.6 & 60.8 &	69.6 & 33.2  \\ 
   2 & $+$prompt,$+$MLM                            & 56.4 & 74.0 & 81.8 & 43.6 & 82.5 & 95.5 & 97.5 & 76.9 & 40.2 & 62.4 &	68.6 & 34.1   \\ 
   3 & $+$prompt,$+$MAP                            & 56.6 & 75.8 & 83.9 & 43.1 & 82.5 & 95.5 & 98.0 & 77.4 & 40.5 & 63.5 &	71.7 & 35.1  \\ 
   4 & $+$prompt,$+$MAP,$+$$\mathcal{L}_{IoU}$     & \textbf{57.0} & \textbf{77.1} & \textbf{83.9} & \textbf{44.4} & \textbf{83.0} & \textbf{96.0} & \textbf{98.0} & \textbf{78.1} & \textbf{42.5} & \textbf{63.7} & \textbf{72.1} & \textbf{36.3}  \\  \hline
    \end{tabular}}
    \vspace{0.1mm}
    \caption{Ablation study of AIMA on Market-1501 Attribute, PETA and PA100K.}
    \vspace{-2mm}
    \label{tab:ablation results}
\end{table*}

\textbf{Ablations on proposed components.} The effectiveness of the prompt template in Sec.~\ref{sec:text_encoder} can be demonstrated by the results of No.0 $vs.$ No.1. The Rank-1 accuracy on the three datasets increased by 2.9\%, 5.0\%, and 1.9\% respectively, and the mAP increased by 3.1\%, 6.9\%, and 1.0\% respectively. Simply converting the given attribute descriptions into corresponding textual sentence can improve performance. This is because sentences contain richer contextual information, which helps the model better understand and match image details.

The effectiveness of the MAP module can be demonstrated by the results of No.1 $vs.$ No.3 and No.2 $vs.$ No.3. The Rank-1 accuracy on the three datasets increased by 6.2\%, 4.0\%, and 2.9\% respectively, and the mAP increased by 3.3\%, 2.5\%, and 1.9\% respectively. This indicates that using MAP allows the model to learn more discriminative global features. Additionally, the results of MAP are consistently better than those of MLM, indicating that it helps the model better understand key attributes in the text while promoting alignment between local image regions and attributes.

The effectiveness of the A-IoU IMC loss can be verified through the results of No.3 $vs.$ No.4. After applying $\mathcal{L}_{IoU}$, the maximum improvement in Rank-1 accuracy reaches 2.0\%, and the maximum boost in mAP is 1.3\%. This demonstrates that $\mathcal{L}_{IoU}$ enables a better semantic arrangement of texts with different categories in the embedding space, thereby enhancing the accuracy of retrieval.
\begin{figure*}[htbp]
    \centering
    \includegraphics[width=\textwidth]{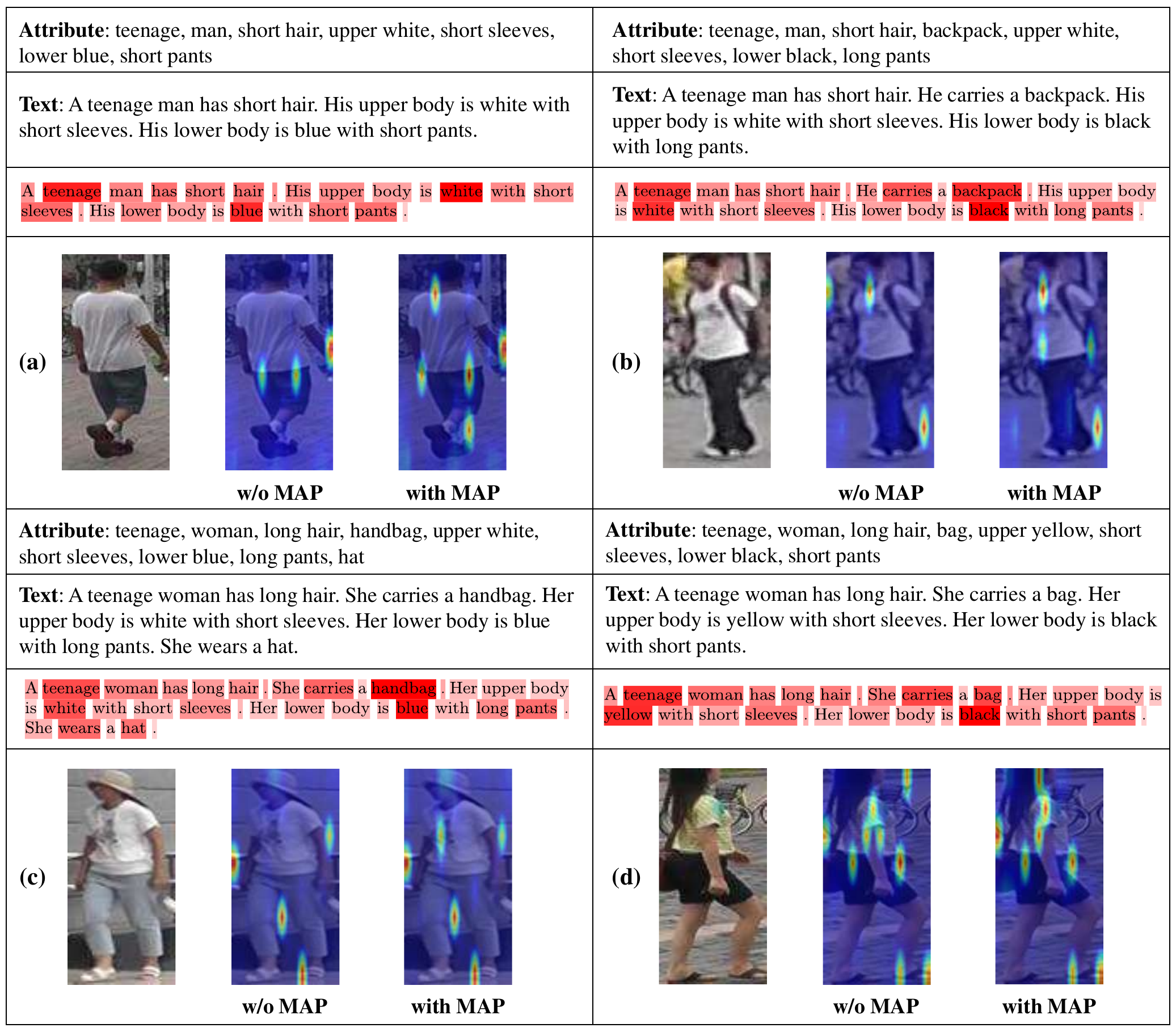}
    \caption{Visualization results of attention weights on Market-1501 Attribute.}
    \label{fig4:attention_weights}
    \vspace{-2mm}
\end{figure*}

\textbf{Further analysis of the MAP module.} To further analyze the role of the proposed MAP, we employ the Attention Rollout~\cite{attention-rollout} method to visualize attention weights for select data samples from Market-1501. As depicted in Fig.~\ref{fig4:attention_weights}, darker colors signify higher model attention. In (a) and (c), the model with the MAP module can focus on more details in the image, which is beneficial for improving discriminative ability. In (b) and (d), the model without MAP focuses on background information unrelated to pedestrians, which is detrimental to the alignment between visual and textual information. Adding MAP allows for implicit alignment between image details and key textual attributes, thereby avoiding the introduction of noise.
\begin{figure*}[htbp]
    \centering
    \includegraphics[width=\textwidth]{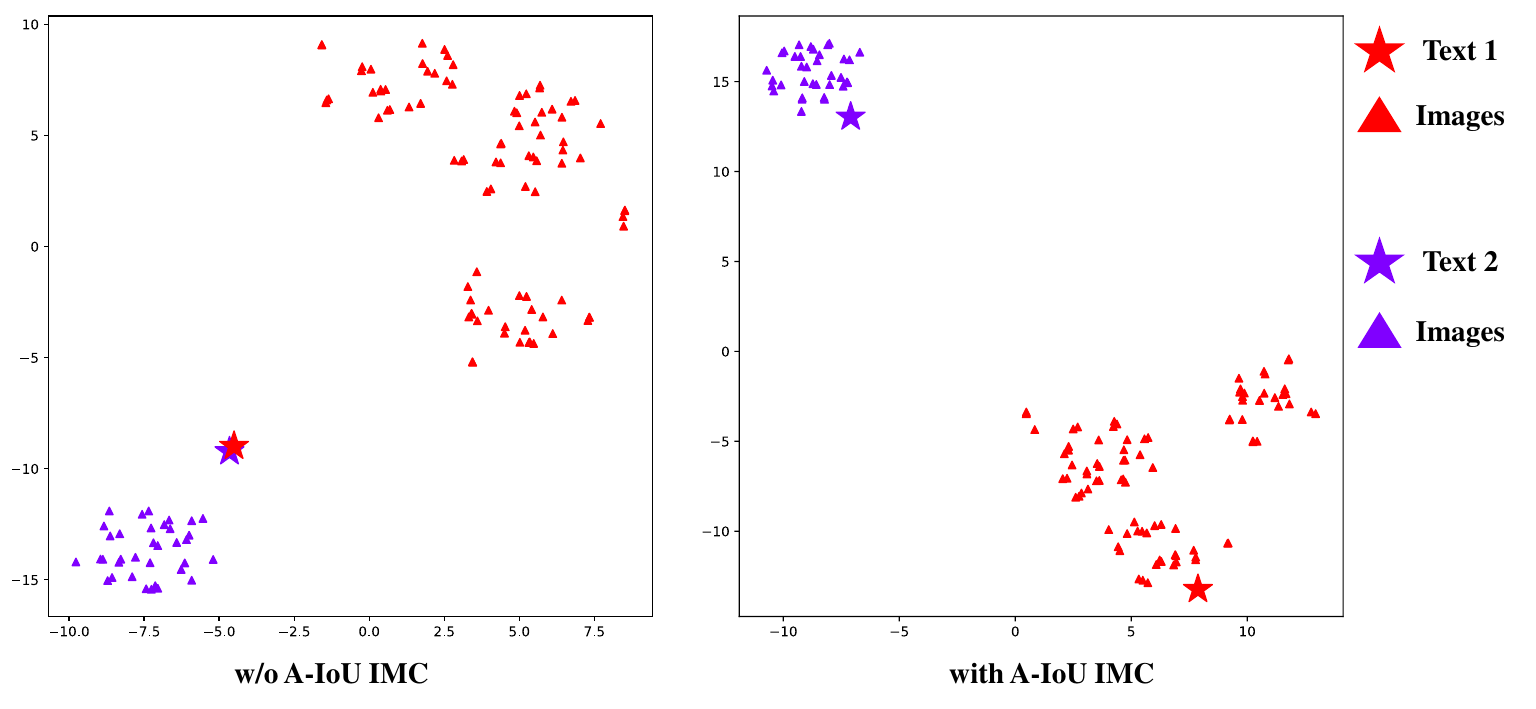}
    \caption{t-SNE visualization of a part of the joint embedding space on Market-1501 Attribute. Stars and triangles indicate embedding vectors of texts and their associated images, respectively, and their colors mean their categories.}
    \label{fig5:embedding_vector}
    \vspace{-2mm}
\end{figure*}

\textbf{Further analysis of the A-IoU IMC loss.} To further analyze the proposed A-IoU IMC loss, we visualize the positions of text queries and images in the embedding space. In the left image of Fig.~\ref{fig5:embedding_vector}, due to the lack of certain constraints, two similar text representations are closer in the embedding space. This results in text queries represented in red being closer to images represented in purple, leading to incorrect retrieval results. In contrast, in the right image of Fig.~\ref{fig5:embedding_vector}, different text representations are constrained based on differences in attribute words, causing them to be separated by a certain distance in the embedding space, thus obtaining correct retrieval results. This further demonstrates that the proposed loss can adaptively separate text representations of different classes in the embedding space.

\begin{figure*}[htbp]
    \centering
    \includegraphics[width=120mm]{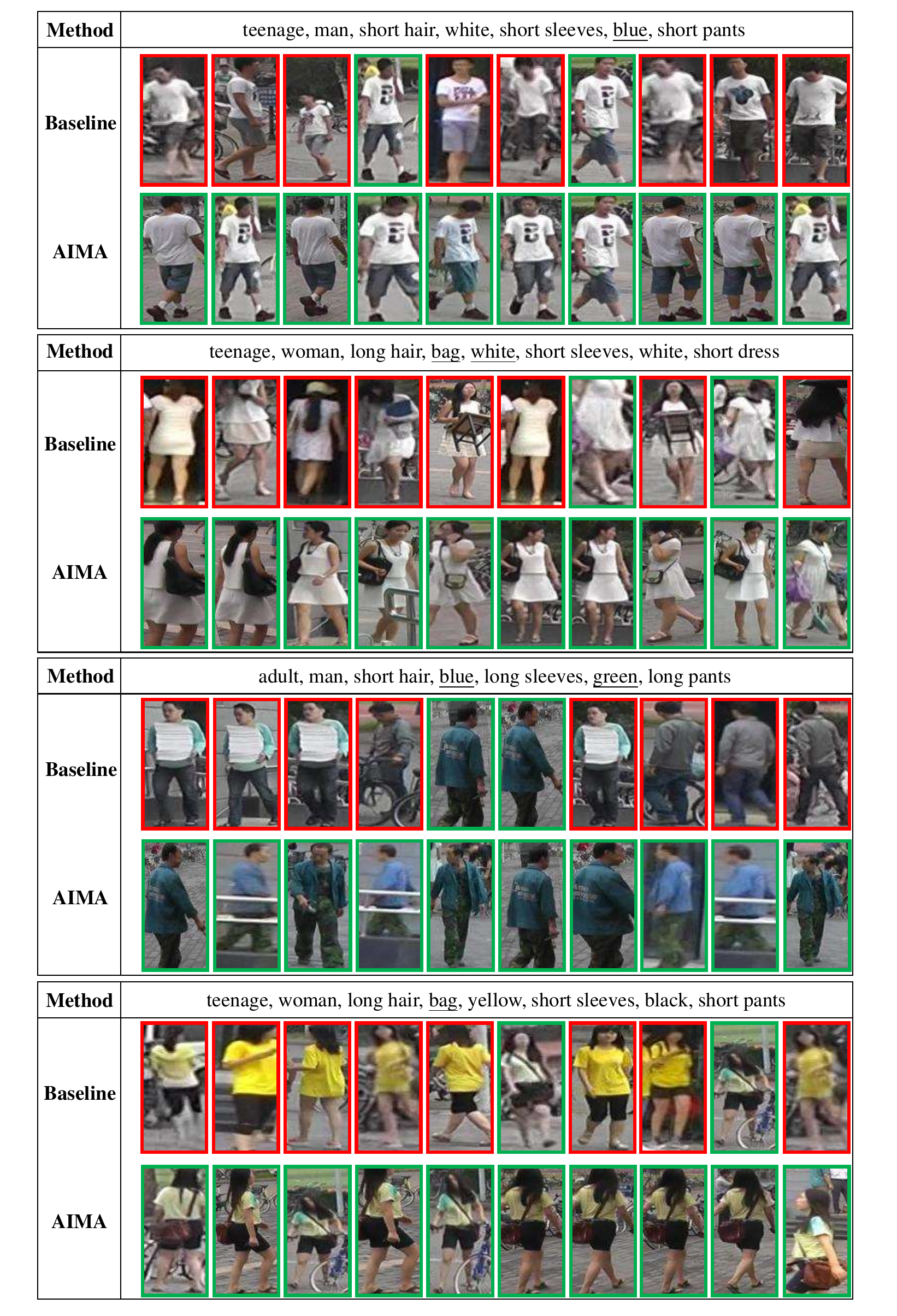}
    \caption{Comparisons of the retrieval results between Baseline and our AIMA on Market-1501 Attribute. True/false image matches are indicated by green/red boxes. Images are sorted from left to right according to their ranks.}
    \label{fig6:visualize_results}
    \vspace{-2mm}
\end{figure*}

\subsection{Qualitative Results}

We visualize some retrieval results and compare them with the Baseline. As shown in Fig.~\ref{fig6:visualize_results}, our AIMA is able to obtain more correct retrieval results. The Baseline fails to align the local images corresponding to the key attribute words in the sentence, resulting in incorrect retrieval results. This is mainly due to the following reasons: Firstly, the text input of the Baseline is composed of multiple attribute words, lacking contextual connections and failing to form an effective holistic representation. The prompt template in AIMA can effectively convert attributes into texts, which are more structured and beneficial for the model to better understand and represent. Additionally, the Baseline lacks modeling of fine-grained information interaction between modalities, making it unable to learn discriminative features. In contrast, AIMA implicitly achieves fine-grained alignment of information through the masked attribute prediction task, enhancing the discriminative power of the model. Lastly, AIMA further improves the accuracy of retrieval by controlling the distribution of corresponding texts through the differences between attributes.

\section{Conclusion}

In this paper, we propose the Attribute-Aware Implicit Modality Alignment (AIMA) framework for text attribute person search. We first introduce the CLIP model as the backbone and design prompt templates to enhance the model's understanding capability. Additionally, we design a Masked Attribute Prediction (MAP) module to achieve implicit local relationship alignment between images and text. This facilitates the model in learning more discriminative global representations. Furthermore, we propose an Attribute-IoU Guided Intra-Modal Contrastive (A-IoU IMC) loss, which enables better semantic arrangement of different categories of text in the embedding space. Extensive experiments demonstrate the superiority and effectiveness of our proposed AIMA. In the future, we aim to further explore more effective ways to apply CLIP to text attribute person search.





\bibliographystyle{elsarticle-num}


\end{document}